\def\ReaderLayout{1}
\documentclass{article}
\usepackage{iclr2027_conference,times}

\usepackage{amsmath,amsfonts,bm}

\def\eqref#1{equation~\ref{#1}}

\def\1{\bm{1}}

\DeclareMathAlphabet{\mathsfit}{\encodingdefault}{\sfdefault}{m}{sl}
\SetMathAlphabet{\mathsfit}{bold}{\encodingdefault}{\sfdefault}{bx}{n}

\usepackage{hyperref,url,graphicx,amsmath,amssymb,booktabs,longtable}
\usepackage{placeins,xcolor,tikz,multirow,array,tabularx,microtype}
\hypersetup{hidelinks}
\definecolor{TwinBlue}{HTML}{204A75}
\usetikzlibrary{arrows.meta,positioning}
\title{Small Frequency Corrections Can Change What Survives KV Cache Compression}
\author{
Hong Chen$^{1}$,
Yudong Zeng$^{2}$,
Yongwei Huang$^{2}$,
Zuhao Ouyang$^{2}$,
Dongnan Zheng$^{2}$,\\
\textbf{Junyan Zhang$^{1}$,
Yubo Gao$^{1}$,
Xuming Hu$^{1,*}$} \\ \\
$^{1}$The Hong Kong University of Science and Technology (Guangzhou), Guangzhou, China \\
$^{2}$Bosum Institute of Management Science, Shenzhen, China \\
$^{*}$Corresponding author
}
\iclrfinalcopy
\newcommand{\Twin}{TwinKV}
\newcommand{\ind}{\mathbb{I}}
\newcommand{\AIStatement}{\section*{AI Use Statement}
We used Generative AI for minor language polishing, grammar checking, and readability improvements of author-written text. These tools did not propose scientific ideas, experimental designs, results,
references, or new technical claims. All suggestions were reviewed, accepted, rejected, or rewritten by the authors. The authors take full responsibility for the final content, including any remaining errors.}
\newcommand{\ReproStatement}{\subsection*{Reproducibility statement}
Section~\ref{sec:method} specifies the algorithm and fixed parameters, and Section~\ref{sec:setup} defines the datasets and evaluation metrics. Appendix~\ref{sec:repro} provides implementation details, baseline configurations, and the budget-conservation proof. Appendix~\ref{sec:cohort-scoring} records data processing, scoring, and the software environment; Appendix~\ref{sec:implementation-checks} describes implementation checks and retained artifacts. The efficiency protocol is detailed in Appendix~\ref{sec:efficiency-details}. Appendix~\ref{sec:new-confirmation} details the component experiments and resampling. Recorded sample identifiers, predictions, cache witnesses, and source hashes support input matching and independent score verification.}
\newif\ifreaderlayout
\ifdefined\ReaderLayout\readerlayouttrue\else\readerlayoutfalse\fi
\begin{document}
\maketitle

\begin{abstract}
Compressing a key-value cache before its next question is known requires choosing what to retain without knowing which evidence will matter. Value energy measures entry strength but does not distinguish isolated keys from those with many similar neighbors. We introduce TwinKV, a training-free method that discounts value energy by nonlocal post-RoPE key frequency. Prefix attention allocates head capacities, while retained entries preserve their original keys and values under an exact storage budget. Across four language models, TwinKV exceeds five evaluated compressed baselines in mean score on LongBench, LooGLE, and RULER at 50\% KV removal. Component controls isolate the frequency contribution. On Llama-3.2-1B RULER at 75\% removal, normalized frequency weights average 0.95, yet change 7\% of nonprotected retained positions and improve value-only retention by about 5.5 points under both uniform and adapted capacities. Permuting the weights within heads weakens this gain. These results show that modest frequency corrections can change retention and answering outcomes, with effects that depend on the model and task.
\end{abstract}

\section{Introduction}
A document can repeat a standard procedure many times and record an exception only once. Its next reader might ask about either. If the document's KV cache must be compressed before that question arrives, the system has to decide what to retain without knowing which answer will matter.

Existing retention policies draw on different signals. StreamingLLM protects attention sinks and a recent window~\citep{xiao2024efficient}; H2O retains heavy hitters alongside recent tokens~\citep{zhang2023h2o}; SnapKV estimates useful positions from a prompt-end observation window~\citep{li2024snapkv}. PyramidKV further changes how capacity is distributed across layers~\citep{cai2024pyramidkv}. These policies make two linked decisions: how much space each part of the cache receives, and which entries occupy that space. Separating the decisions lets us examine the retention signal itself.

KeyDiff uses key distinctiveness~\citep{park2026keydiff}, while NestedKV combines deviations from key anchors at several temporal scales~\citep{chen2026nestedkv}. Value-aware pruning shows that value norms carry information beyond attention scores~\citep{guo2024attention}. Does the number of similar nonlocal keys add useful information to value energy, once allocation and protection are held fixed?

Value energy measures entry strength; nonlocal key frequency measures how often similar keys occur elsewhere in the prefix. An energy-based ranking treats equally strong entries alike even when their frequencies differ. \Twin\ discounts value energy by key frequency and selects original KV entries within a fixed budget. Observed prefix attention sets head capacities through an adaptation inspired by Ada-KV~\citep{feng2026ada}.

The complete rule improves the four-model mean on each of the three benchmarks relative to the five stated baselines at 50\% removal.

We examine frequency weighting under both uniform and adapted head capacities by crossing value-only versus Twin scoring with the two capacity rules. The comparison identifies a clear frequency contribution on Llama-3.2-1B RULER at 75\% removal, under both capacity rules. A second intervention permutes frequency weights within each head, preserving their distribution and the cache budget. The real correspondence outperforms all three predeclared permutations in that setting.

Our contributions are a frequency-discounted retention rule under exact protected budgets, a three-benchmark evaluation, and component experiments connecting measured counts, changed selections, and answer scores.

\section{What Can Be Decided Before the Question?}
\label{sec:motivation}
A compressor maps a context $C$ and its full cache $\mathcal K(C)$ to a retained prefix $\widehat{\mathcal K}(C)$ before observing a future question $q$. Selection applies the same static rule to every prefix, using the prefix alone. Each later question reads this fixed cache with its own suffix state.

\subsection{An empirical reason to combine the signals}
\label{sec:empirical-motivation}
\begin{figure}[t]
\centering
\includegraphics[width=\linewidth]{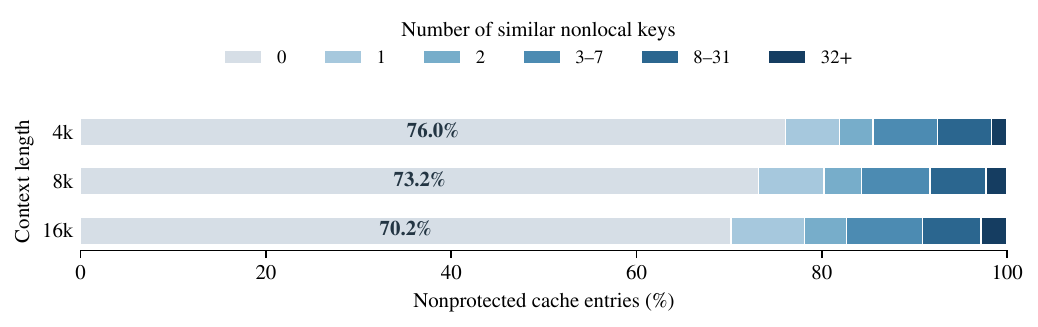}
\caption{\textbf{Post-RoPE frequency distributions.} Llama-3.2-1B on RULER; context-averaged proportions of nonprotected entries. Gray indicates zero nonlocal neighbors.}
\label{fig:frequency-distribution}
\end{figure}

The distribution of real key counts motivates combining frequency with value energy. Figure~\ref{fig:frequency-distribution} measures the actual post-RoPE counts on RULER contexts with Llama-3.2-1B. Across the three lengths, 73.13\% of nonprotected entries have no similar nonlocal key. Such entries all receive the same maximal rarity weight. Value energy orders this large set of equally rare entries. Meanwhile, the remaining 26.87\% receive nontrivial discounts, allowing repetition to change an energy-based decision.

\begin{figure}[t]
\centering
\includegraphics[width=\linewidth]{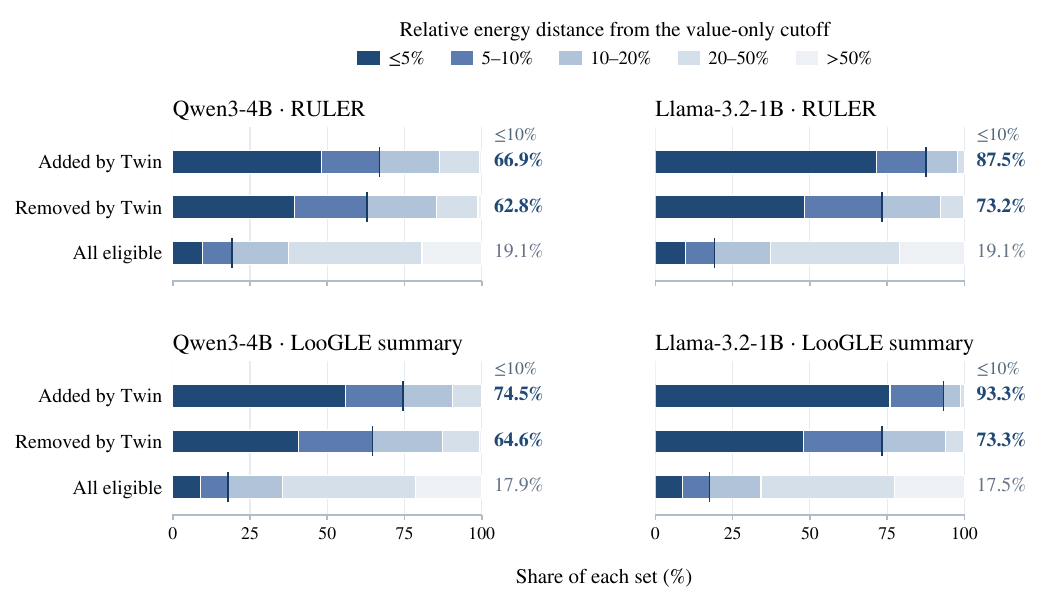}
\caption{\textbf{Energy margins at the selection boundary.} At 75\% removal, segments partition each set by distance from its head's value-only cutoff. Labels give the share within 10\%. Percentages are population-weighted within each context and then averaged.}
\label{fig:selection-boundary}
\end{figure}

Figure~\ref{fig:selection-boundary} locates the entries exchanged by the frequency correction. We compare positions added to and removed from the value-only retained set with all nonprotected candidates, using the same adapted head capacities. Across both models and task settings, the changed sets are concentrated near this boundary. The energy ranking therefore presents many close decisions for the frequency correction to resolve. Appendix~\ref{sec:boundary-distribution} specifies the measurement.

\section{TwinKV: Frequency-Discounted Retention}
\label{sec:method}
\begin{figure}[t]
\centering
\includegraphics[width=\linewidth]{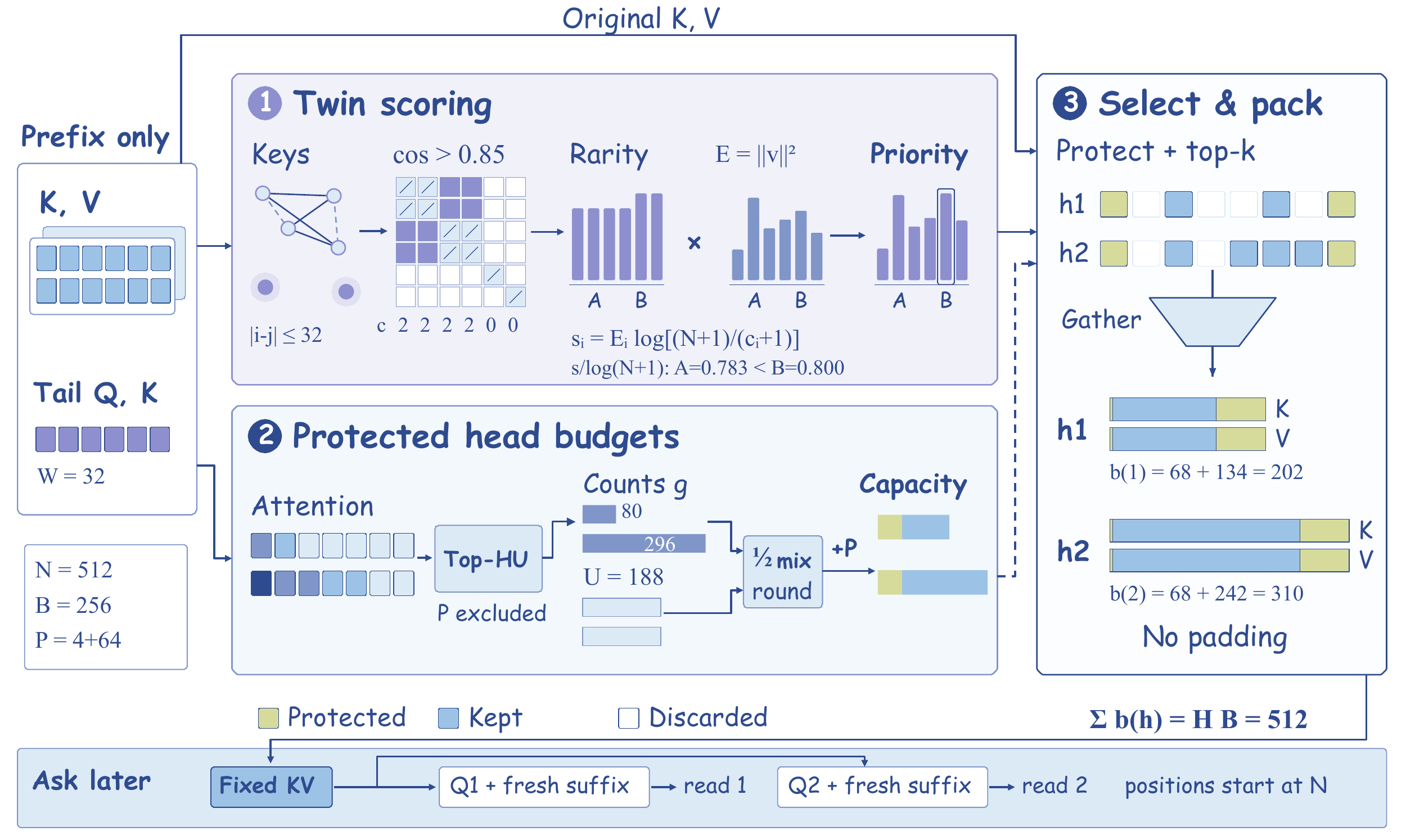}
\caption{\textbf{TwinKV per layer.} Frequency-weighted value scoring, protected head budgets, and original-KV packing. The bottom strip shows later reads with independent suffixes. Counts, ranking examples, and capacities are illustrative (Appendix~\ref{sec:repro}).}
\label{fig:method}
\end{figure}

Figure~\ref{fig:method} separates the score used to choose positions from the capacity assigned to each head.
\subsection{Prefix-only compression under an exact budget}
Consider one layer with $H$ KV heads and a prefix of $N$ positions. For head $h$, let $k_{h,i},v_{h,i}\in\mathbb{R}^{d}$ denote the cached key and value. Keys are the actual \emph{post-RoPE} keys used by the model. We define $r$ as the fraction of prefix positions removed and set
\begin{equation}
 B=\lfloor N(1-r)\rfloor, \qquad \sum_{h=1}^{H}b_h=HB,
 \label{eq:budget}
\end{equation}
where $b_h$ is the retained capacity of head $h$. The target is enforced separately for each layer. The tested ratios $r\in\{0.25,0.50,0.75\}$ retain approximately $75\%,50\%,25\%$ of the original prefix KV storage.

We reserve the first $s=\min(4,B)$ and the last $w=\min(64,B-s)$ positions. Write their union as $\mathcal P$, with $P=s+w$, and let $U=B-P$. The same protected positions are retained in every head; they do not add capacity beyond Equation~\ref{eq:budget}. On short prefixes, this order gives sinks priority and uses the remaining slots for recent positions. Appendix~\ref{sec:repro} covers boundary cases.

\subsection{Value-weighted nonlocal key rarity}
\label{sec:score}
We normalize each key in FP32 as $\hat{k}_{h,i}=k_{h,i}/\max(\|k_{h,i}\|_2,10^{-12})$. Its nonlocal repetition count is
\begin{equation}
 c_{h,i}=\sum_{j=1}^{N}\ind\!\left[|i-j|>\ell\right]
 \ind\!\left[\hat{k}_{h,i}^{\top}\hat{k}_{h,j}>\tau\right],
 \qquad \ell=32,\quad \tau=0.85.
 \label{eq:count}
\end{equation}
The local exclusion removes immediately adjacent positions from the count. Because keys are post-RoPE, their similarity depends on position as well as content.

The selection score is
\begin{equation}
 e_{h,i}=\|v_{h,i}\|_2^2,\qquad
 s_{h,i}=e_{h,i}\log\frac{N+1}{c_{h,i}+1}.
 \label{eq:score}
\end{equation}
Dividing by the positive constant $\log(N+1)$ gives the same within-head ranking:
\begin{equation}
 \frac{s_{h,i}}{\log(N+1)}=e_{h,i}\,d_{h,i},\qquad
 d_{h,i}=1-\frac{\log(c_{h,i}+1)}{\log(N+1)}.
 \label{eq:discount}
\end{equation}
An entry with no nonlocal neighbor has $d_{h,i}=1$; more neighbors reduce the multiplier. Equal counts recover the value-only ranking. We evaluate Equation~\ref{eq:score} in FP32.

Counts measure neighborhoods in post-RoPE key space. We compute them once over the original prefix and use them throughout selection. Protected positions also contribute to the counts, so each entry's frequency reflects the full prefix. Exact score ties use descending value energy, then ascending position.

\textbf{Why a small correction can change retention.} Selection depends on score margins near the capacity boundary, not on the average multiplier alone. For two unprotected entries in one head with $e_i>e_j>0$, frequency weighting reverses their order when $1<e_i/e_j<d_j/d_i$. A modest relative difference between the discounts can therefore matter when the energy margin is small. Conversely, an entry with sufficiently large energy can survive a substantial frequency penalty. This is the intended interaction between strength and repetition: neither term independently determines the retained set. The constructed A/B pair in Figure~\ref{fig:method} illustrates the reversal: energies $0.95/0.80$ become normalized scores $0.783/0.800$ after frequency weighting.

The mean also hides heterogeneity: many zero-count entries can keep the average weight near one even when a minority receives strong penalties.

\subsection{Head allocation adapted to rarity-weighted selection}
\label{sec:allocation}
Scaling a head's values changes score magnitudes without changing their order. We therefore allocate cross-head capacity separately from within-head scoring. Inspired by Ada-KV, we allocate only the $HU$ unprotected slots, using the final $W=32$ observed prefix queries as the capacity signal. The $G$ query heads sharing a KV head are averaged together; causal attention from these 32 positions is max-pooled with width seven. After excluding protected positions, the resulting FP32 scores $z_{h,i}$ determine the head capacities. Appendix~\ref{sec:allocation-details} specifies attention reconstruction and boundary handling.

We adapt Ada-KV's global-count rule~\citep{feng2026ada} to this residual capacity: take the largest $HU$ scores from unprotected head--position pairs, after removing the protected set from eligibility. Let $g_h$ count how many belong to head $h$, so $\sum_h g_h=HU$. Equal scores use a stable order: head index followed by token position. We mix this allocation equally with the uniform unprotected budget:
\begin{equation}
 \widetilde u_h=\frac{g_h+U}{2},\qquad
 u_h=\lfloor\widetilde u_h\rfloor+\delta_h,\qquad b_h=P+u_h.
 \label{eq:quota}
\end{equation}
Largest-remainder correction assigns the $\delta_h$ increments deterministically (Appendix~\ref{sec:allocation-details}). The half-uniform floor reserves slots even in heads receiving little prefix-tail attention; Twin's ranking fills them.

\textbf{Budget conservation.} Equation~\ref{eq:quota} gives $\sum_h b_h=HB$ and $b_h\ge P+\lfloor U/2\rfloor$: rounding redistributes slots without changing their total. The proof appears in Appendix~\ref{sec:repro}.

\subsection{Selection and physical cache layout}
\label{sec:packing}
Retain $\mathcal P$ and the top $b_h-P$ unprotected positions by Equation~\ref{eq:score}, then gather their original K/V in sequence order. Contiguous heads with equal quotas share a tensor; unequal quotas are not padded. The prefix payload is exactly $2dHB$ elements per layer, excluding weights, temporary buffers, metadata, and suffix caches. Native SDPA reads each head group. Query positions continue from the original prefix length $N$; keys are not rephased and values are unchanged.

\subsection{The cost of deciding what to retain}
Nonlocal counting costs $O(HN^2d)$ per layer. Processing $C=512$ query keys at a time bounds the similarity workspace by $O(HCN)$. Observation statistics cost $O(HGWN d)$, with additional sorting for allocation and selection. Section~\ref{sec:efficiency} measures construction, reading, and peak memory separately.

\section{Evaluating the Retained Context}
\label{sec:setup}
\textbf{Models and baselines.} We use Qwen3-4B and Qwen3-8B~\citep{yang2025qwen3}, Llama-3.2-1B-Instruct and Llama-3.1-8B-Instruct~\citep{grattafiori2024llama} without weight updates. The five compressed baselines are StreamingLLM, PyramidKV, SnapKV, KeyDiff~\citep{park2026keydiff}, and H2O~\citep{zhang2023h2o}. The H2O column follows the prefix-only observation rule specified in Appendix~\ref{sec:h2o-implementation}. FullKV retains the complete capped prefix and is a separate reference. All methods compress the prefix only; generated tokens are not evicted. \Twin\ uses exactly the parameters in Section~\ref{sec:method} for every model, benchmark, and ratio. Appendix~\ref{sec:repro} records implementation details.

\textbf{Datasets.} We evaluate on LongBench~\citep{bai2024longbench}, LooGLE~\citep{li2024loogle}, and RULER~\citep{hsieh2024ruler}. LongBench combines question answering, summarization, and code tasks drawn from natural contexts. LooGLE examines long-document understanding through short- and long-dependency questions and summarization. RULER offers controlled tests of retrieval, aggregation, and reasoning as context length grows. Together, they cover both locating evidence and using information distributed throughout a prefix. All methods use the same evaluation inputs; sampling and preprocessing are specified in Appendix~\ref{sec:cohort-scoring}.

\textbf{Evaluation protocol.} Methods share tokenized contexts and questions, answer prefixes, truncation, generation limits, and greedy stopping rules. FullKV is evaluated once and reused as the uncompressed reference. We keep each benchmark's native primary metric: LongBench averages task-specific scores, LooGLE averages its primary split scores, and RULER averages task scores within each length before aggregating across lengths. The reported scale is 0--100. Appendix~\ref{sec:cohort-scoring} gives the exact scoring and aggregation rules.

\textbf{Component controls.} We cross value-only and Twin scoring with uniform and adapted head capacities. Within each model--prefix--ratio comparison, protection, model weights, the reader, and total KV bytes are fixed; retained entries keep their original K/V. Score comparisons reuse the per-prefix head capacities, while capacity comparisons keep the scoring rule fixed. Frequency permutations and matched replacements then test which changes in the retained set matter for answering. Appendix~\ref{sec:new-confirmation} gives the component-experiment settings and statistical analysis.

\section{What Does the Retained Context Support?}
\label{sec:results}
\subsection{Benchmark quality across memory budgets}
At $r=0.50$, \Twin\ achieves four-model means of 33.63, 36.58, and 80.81 on LongBench, LooGLE, and RULER, respectively (Table~\ref{tab:main-comparison}). These exceed the strongest compressed baseline means by 1.48, 0.63, and 4.86 points. All four models lead on LongBench and LooGLE; Table~\ref{tab:main-comparison} gives the corresponding model-level RULER results.

\begin{figure}[t]
\centering
\includegraphics[width=\linewidth]{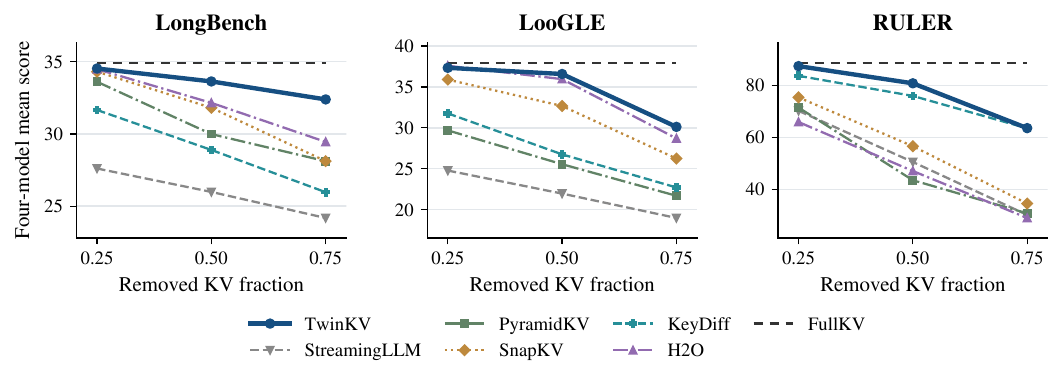}
\caption{\textbf{Quality across removal ratios.} Four-model means on the three benchmarks. FullKV is an uncompressed reference; panels use separate score ranges.}
\label{fig:budget-curves}
\end{figure}

Figure~\ref{fig:budget-curves} shows how quality changes across the three removal ratios. \Twin\ has the highest four-model mean among the compressed methods on all three benchmarks at $r=0.50$ and $0.75$. At $r=0.75$, its LongBench and LooGLE means are 32.38 and 30.12, compared with the strongest baseline means of 29.45 and 28.71. RULER is closer: 63.55 versus KeyDiff's 63.49.

The benchmark means weight models equally. FullKV provides a reference at the original cache size, while the compressed methods share the stated removal ratio.
\begin{table}[t]
\centering
\caption{\textbf{Primary results at 50\% KV removal.} Bold: best compressed score. Italic FullKV: uncompressed reference.}
\label{tab:main-comparison}
\begingroup\footnotesize
\setlength{\tabcolsep}{2pt}
\renewcommand{\arraystretch}{1.12}
\begin{tabular*}{\linewidth}{@{\extracolsep{\fill}}llrrrrrrr@{}}
\toprule
\multirow{2}{*}{Model} & \multirow{2}{*}{Benchmark} & \multicolumn{5}{c}{Compressed baselines} & \multirow{2}{*}{\textbf{TwinKV}} & \multirow{2}{*}{\emph{FullKV}} \\
\cmidrule(lr){3-7}
 & & Stream. & Pyramid & SnapKV & KeyDiff & H2O & & \\
\midrule
\multirow{3}{*}{Qwen3-4B} & LongBench & 26.92 & 32.55 & 35.61 & 26.64 & 35.62 & \textbf{37.20} & \emph{38.37} \\
 & LooGLE & 23.85 & 28.32 & 37.00 & 24.39 & 41.32 & \textbf{41.70} & \emph{42.83} \\
 & RULER & 53.59 & 40.39 & 60.15 & 74.00 & 50.38 & \textbf{85.52} & \emph{92.13} \\
\addlinespace[4pt]
\multirow{3}{*}{Llama-3.2-1B} & LongBench & 20.30 & 20.99 & 21.80 & 21.71 & 21.56 & \textbf{21.91} & \emph{24.58} \\
 & LooGLE & 16.53 & 16.49 & 16.89 & 16.77 & 19.89 & \textbf{19.97} & \emph{21.44} \\
 & RULER & 39.66 & 22.54 & 35.36 & \textbf{65.33} & 29.01 & 61.56 & \emph{71.65} \\
\addlinespace[4pt]
\multirow{3}{*}{Qwen3-8B} & LongBench & 27.74 & 35.68 & 35.80 & 30.90 & 37.10 & \textbf{38.80} & \emph{39.22} \\
 & LooGLE & 23.62 & 31.78 & 39.00 & 27.49 & 42.82 & \textbf{43.29} & \emph{44.63} \\
 & RULER & 56.36 & 43.70 & 59.59 & 78.00 & 53.06 & \textbf{88.82} & \emph{94.08} \\
\addlinespace[4pt]
\multirow{3}{*}{Llama-3.1-8B} & LongBench & 28.98 & 30.72 & 34.02 & 36.30 & 34.29 & \textbf{36.60} & \emph{37.41} \\
 & LooGLE & 23.86 & 25.63 & 37.81 & 38.38 & 39.79 & \textbf{41.37} & \emph{42.73} \\
 & RULER & 52.39 & 67.25 & 71.17 & 86.50 & 55.66 & \textbf{87.34} & \emph{96.45} \\
\bottomrule
\end{tabular*}
\endgroup
\end{table}

\subsection{Behavior as the context grows}
\label{sec:length-evidence}
\begin{figure}[t]
\centering
\includegraphics[width=\linewidth]{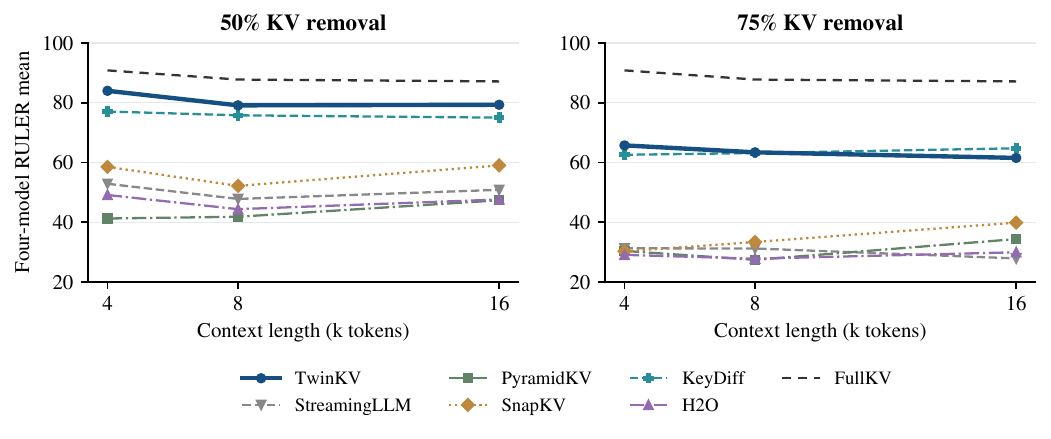}
\caption{\textbf{RULER quality across context lengths.} Means over four models and all 13 tasks at each length; FullKV is an uncompressed reference.}
\label{fig:length-evidence}
\end{figure}

At 50\% removal, \Twin's four-model RULER means exceed the strongest compressed comparator, KeyDiff, by 6.93, 3.33, and 4.31 points at 4k, 8k, and 16k (Figure~\ref{fig:length-evidence}). The advantage at this budget extends across the full tested length range. Figure~\ref{fig:length-evidence} also reports 75\% removal, placing the two budgets on the same length axis. Every point averages all 13 tasks.

\subsection{Ablating the two factors of the Twin score}
\label{sec:confirmed-effects}
\begin{table}[t]
\centering
\caption{\textbf{Core-score ablation.} Llama-3.2-1B, RULER, 50\% removal. Avg. equally weights the three lengths; each length includes all 13 tasks.}
\label{tab:core-ablation}
\small
\setlength{\tabcolsep}{7pt}
\begin{tabular*}{\linewidth}{@{\extracolsep{\fill}}lccrrrr@{}}
\toprule
\multirow{2}{*}{Score variant} & \multicolumn{2}{c}{Score factors} & \multicolumn{4}{c}{RULER score} \\
\cmidrule(lr){2-3}\cmidrule(lr){4-7}
 & Frequency & Value & 4k & 8k & 16k & Avg. \\
\midrule
\textbf{TwinKV} & $\checkmark$ & $\checkmark$ & \textbf{65.72} & \textbf{60.90} & \textbf{58.05} & \textbf{61.56} \\
Without frequency & --- & $\checkmark$ & 63.28 & 59.06 & 52.90 & 58.41 \\
Without value & $\checkmark$ & --- & 49.72 & 47.51 & 53.14 & 50.12 \\
\bottomrule
\end{tabular*}
\end{table}

Table~\ref{tab:core-ablation} isolates frequency discounting and value energy on Llama-3.2-1B RULER at $r=0.50$. The variants share head capacities, protected positions, and the reader (Section~\ref{sec:setup}). Removing frequency gives value-only selection; removing value gives rarity-only selection, with exact count ties resolved by original position rather than by value.

The full score improves on value-only and rarity-only by 3.14 and 11.43 points, respectively. The complete score also exceeds both controls at every evaluated length. Both factors contribute in this comparison: energy orders the many entries with equal counts, and the frequency discount adjusts their priority when counts differ.

\subsection{Frequency weighting under different head budgets}
\begin{table}[t]
\centering
\caption{\textbf{Frequency correspondence.} Llama-3.2-1B, RULER, 75\% removal. Permuted frequency averages three fixed seeds; Avg. weights lengths equally.}
\label{tab:correspondence}
\small
\setlength{\tabcolsep}{9pt}
\begin{tabular*}{\linewidth}{@{\extracolsep{\fill}}lrrrr@{}}
\toprule
\multirow{2}{*}{Selection score} & \multicolumn{4}{c}{RULER score} \\
\cmidrule(lr){2-5}
 & 4k & 8k & 16k & Avg. \\
\midrule
Value-only & 33.67 & 31.78 & 30.84 & 32.10 \\
Permuted frequency & 33.62 & 29.12 & 27.94 & 30.22 \\
\textbf{TwinKV} & \textbf{39.22} & \textbf{35.96} & \textbf{37.72} & \textbf{37.63} \\
\bottomrule
\end{tabular*}
\end{table}

Table~\ref{tab:correspondence} reports the RULER comparison under adapted head capacities. At $r=0.75$ on Llama-3.2-1B, Twin scoring improves value-only selection by 5.54 points under adapted capacity and 5.48 under uniform capacity. Repeating the comparison on another RULER sample, the same Llama-3.2-1B comparison yields gains of 3.62 points with adapted capacity and 3.27 with uniform capacity, again at $r=0.75$.

\subsection{From key counts to frequency--position correspondence}
\label{sec:measured-counts}
\label{sec:frequency-shuffle}
For the RULER inputs in Figure~\ref{fig:frequency-distribution}, the mean normalized discount is 0.9465. Frequency weighting changes 7.12\% of nonprotected retained positions relative to value-only at the same adapted capacities.

We permute nonprotected counts within each layer and KV head, preserving their multiset. The three seeds are fixed in advance and use no future questions. True Twin exceeds their mean at every length (Table~\ref{tab:correspondence}), with an aggregate gap of 7.41 points. The observed frequencies provide a more useful correction here than the same weights assigned to other positions.

Appendix~\ref{sec:new-confirmation} reports the complete model--ratio controls and the permutation protocol. Because a permutation can alter many more retained positions, we also examine replacements matched to the size and location of the actual selection change.

The matched-replacement experiment covers every RULER task and length. For each layer and head, the control swaps exactly as many positions as the change from value-only to Twin. Removed and added entries are sampled within the same joint bins of value-energy rank and position as the corresponding Twin replacements. Three fixed seeds change the assignment.

On this Llama-3.2-1B diagnostic at $r=0.75$, value-only scores 31.13 and Twin scores 37.54. The three matched controls score 32.37, 31.90, and 30.56. Twin exceeds their mean by 5.93 points. Matching the amount, energy range, and location of the changes therefore leaves an advantage for the actual frequency-based selection in this diagnostic. Appendix~\ref{sec:matched-replacements} gives the sampling and matching details.

Displaced entries in this diagnostic average 7.75\% above their head's value-only energy cutoff; promoted entries average 4.39\% below it. The estimates weight samples by group size before averaging contexts (Appendix~\ref{sec:matched-replacements}). These measured margins place the changes near the selection boundary, where frequency corrections can reverse the energy ordering. Figure~\ref{fig:selection-boundary} shows the full distributions: 87.52\% of added entries and 73.24\% of removed entries lie within 10\% of the cutoff, compared with 19.08\% of all eligible candidates.

\subsection{Prefill, decoding, and memory cost}
\label{sec:efficiency}
\begin{figure}[t]
\centering
\includegraphics[width=\linewidth]{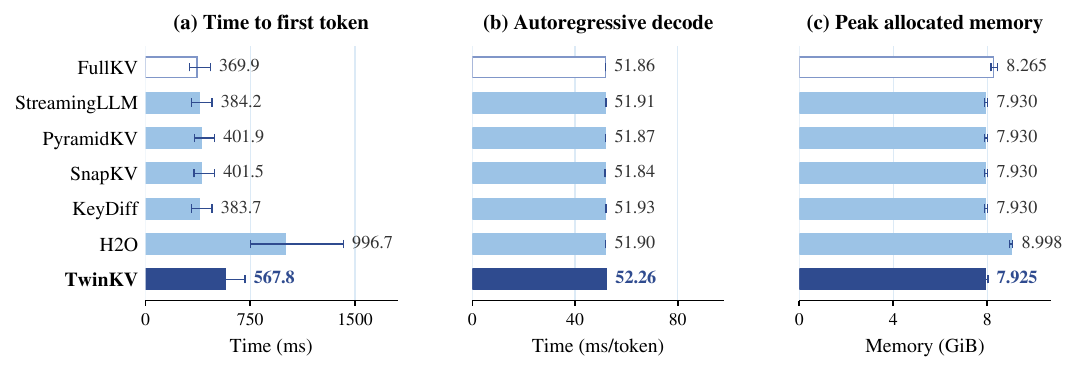}
\caption{\textbf{Stage costs.} Qwen3-4B, RULER-4k, 75\% removal. Bars average context medians; whiskers show their range. Hollow bars denote FullKV.}
\label{fig:efficiency-stages}
\end{figure}
Figure~\ref{fig:efficiency-stages} compares the complete baseline set on Qwen3-4B/RULER-4k at 75\% removal, measured with matched SDPA implementations on one NVIDIA L20. All methods use the same fixed contexts and measurement procedure. TTFT includes context prefill, compression, and question processing through the first token. Decode reports time per generated token after the first-token stage. Peak allocated memory covers construction and fresh-suffix reads, including model weights and temporary tensors. Appendix~\ref{sec:efficiency-details} specifies timing and aggregation.

Twin's 7.925\,GiB peak is close to the lighter compressed baselines' 7.930\,GiB and below FullKV and H2O. Frequency counting adds construction work, while unequal head capacities create additional reader groups.

\section{Limitations}
\textbf{Evaluation scope.} The results concern fixed-prefix reading; multi-fact question coverage, online agent rollouts, and unbounded streaming remain outside this evaluation.

\textbf{Design and implementation.} Key frequency measures geometric repetition. Similar keys may carry different values, and their neighbors may be removed during selection. The method uses fixed similarity and local-exclusion parameters. Its exact budget describes prefix K/V storage; total memory also includes weights, workspaces, and suffix state. Blockwise counting and grouped SDPA incur the costs measured in Section~\ref{sec:efficiency}. The evidence supports a quality--storage tradeoff under this implementation.

\section{Related Work}
\textbf{Attention and recency as retention signals.} StreamingLLM combines attention sinks with a recent window~\citep{xiao2024efficient}. H2O retains heavy hitters alongside recent tokens~\citep{zhang2023h2o}, while SnapKV estimates importance from a prompt-end observation window~\citep{li2024snapkv}. Twin uses observed attention to assign head capacities and key frequency with value energy to choose their contents.

\textbf{Key distinctiveness and repetition.} KeyDiff ranks keys by deviation from a prefix anchor~\citep{park2026keydiff}. NestedKV uses global, block, and recent-window anchors with head-adaptive mixing~\citep{chen2026nestedkv}. Twin instead counts keys within a thresholded nonlocal neighborhood. Keys at similar distances from an anchor can have different neighbor counts, so distinctiveness and frequency need not induce the same ranking. Twin applies the frequency statistic as a discount on value energy, preserving both signals in the selection rule.

\textbf{Value information and semantic units.} VATP combines attention scores with value norms~\citep{guo2024attention}, and AsymKV distinguishes local key homogeneity from value heterogeneity~\citep{cui2026homogeneous}. ChunkKV selects chunks to preserve local structure~\citep{liu2026chunkkv}; R-KV addresses redundancy in reasoning caches~\citep{cai2026r}. Semantic Integrity Matters introduces KVFundaBench to distinguish sparse retrieval from high-density reasoning and proposes ShotKV, which preserves few-shot examples as semantic units~\citep{liu2025semantic}. These methods highlight the importance of both stored content and retention granularity. Twin keeps original entries and uses nonlocal frequency to adjust their individual priorities.

\textbf{Capacity allocation.} PyramidKV distributes memory across layers~\citep{cai2024pyramidkv}, while Ada-KV derives adaptive head budgets from an attention-output objective~\citep{feng2026ada}. Twin adapts the global-count rule and uniform floor to the budget remaining after sink and recent-position protection. GQA averaging and deterministic integer correction preserve the exact layer budget.

\textbf{Memory across dialogue turns.} FlowKV isolates compressed history from earlier turns and compresses the latest completed turn, avoiding repeated compression of older context~\citep{liu2025flowkv}. SONIC learns to represent historical segments with compact Nexus tokens and uses dynamic budget training to support different memory constraints~\citep{chen2026sonic}. Both organize cache management around dialogue structure. Twin studies a complementary decision: choosing original KV entries within a prefix before its future question is available.

\textbf{Precision and serving systems.} KIVI reduces cache precision through asymmetric quantization~\citep{liu2024kivi}; PagedAttention manages memory fragmentation and cache sharing~\citep{kwon2023efficient}. Twin retains original-precision entries and uses a matched SDPA reader. Its payload accounting and stage-cost measurements distinguish the memory saved by selection from the work of constructing and reading the resulting cache.

\section{Conclusion}
TwinKV discounts value energy by nonlocal key frequency and retains original KV entries under exact protected head budgets. In the RULER component experiments, Llama-3.2-1B gains from frequency weighting under both uniform and adapted capacities. An intervention matches the number, energy range, and position of KV replacements; the actual frequency-based selections outperform all three fixed matched controls. Together, the count distributions, score margins, and answering scores identify how small frequency corrections act: they change which entries cross the budget boundary, and the identity of those entries matters for later answers.

\FloatBarrier
\label{main-text-end}
\clearpage
\ifreaderlayout\else
\AIStatement
\ReproStatement
\bibliography{iclr2027_conference}
\bibliographystyle{iclr2027_conference}
\fi
\clearpage
\appendix
\section{Algorithm and Reproducibility Details}
\suppressfloats[t]
\label{sec:repro}
\subsection{Boundary handling and evaluation storage}
The first and last protected ranges do not overlap for the tested prefixes; if $U=0$, allocation is skipped. The shortest tested prefix has 172 tokens: at $r=0.75$, its 43-slot budget contains four sinks and 39 recent positions, rather than assuming a fixed 68-slot protection overhead. The original quality helper temporarily constructs a uniform control cache. It is released before decoding and excluded from retained-payload accounting; that helper does not measure production peak memory. Figure~\ref{fig:method} uses the constructed example $H=2$, $N=512$, $r=0.5$, $P=68$, and $b=(202,310)$; its glyphs are not model measurements. The six displayed keys have counts $(2,2,2,2,0,0)$. Candidates A/B have energies $0.95/0.80$ and normalized scores approximately $0.783/0.800$, illustrating a rank reversal. The lower strip depicts fixed-prefix reuse, not an answer-coverage experiment.

\subsection{Prefix-attention capacity statistics}
\label{sec:allocation-details}
For $G$ query heads sharing KV head $h$, let $q_{h,g,t}$ be the post-RoPE query at prefix position $t\in\{N-W+1,\ldots,N\}$. We reconstruct causal attention to prefix keys:
\begin{equation}
 a_{h,g,t,i}=\begin{cases}
 \displaystyle\frac{\exp(q_{h,g,t}^{\top}k_{h,i}/\sqrt d)}
 {\sum_{j=1}^{t}\exp(q_{h,g,t}^{\top}k_{h,j}/\sqrt d)}, & i\le t,\\
 0, & i>t.
 \end{cases}
 \label{eq:attention}
\end{equation}
We average over the $G$ query heads and $W$ positions, discard the last $W$ key positions from the allocation score array, and apply stride-one max pooling with width seven:
\begin{equation}
 z_{h,i}=\operatorname{MaxPool}_{7}\!\left(
 \frac{1}{GW}\sum_{g,t}a_{h,g,t,i}\right), \quad 1\le i\le N-W.
 \label{eq:pool}
\end{equation}
Pooling uses three implicit negative-infinity entries at each boundary, so padding cannot become a maximum. Protected positions are excluded \emph{after} pooling. Softmax, averaging, and pooling use FP32; we do not subsequently normalize scores within a head. For all tested prefixes with $U>0$, the protected tail includes the observation window, so every unprotected position has an allocation score.

Equal allocation scores use head index, then token position. In Equation~\ref{eq:quota}, $R=HU-\sum_h\lfloor\widetilde u_h\rfloor$ heads receive $\delta_h=1$: the first $R$ with fractional remainder $1/2$. Others receive zero. If all scores are zero, set $g_h=U$; the floor coefficient stays $0.5$ at every removal ratio.

\subsection{Budget proof}
\textbf{Budget conservation.} Equation~\ref{eq:quota} satisfies $\sum_h b_h=HB$ and $b_h\ge P+\lfloor U/2\rfloor$. Indeed, the interpolated quotas sum to $HU$, and the $R$ increments restore exactly the amount removed by integer rounding. Adding $P$ slots to each head yields $H(U+P)=HB$. Moreover, $g_h\le N-P$ and $U\le N-P$, hence the rounded unprotected quota cannot exceed $N-P$. The guarantee concerns exact prefix storage and the minimum protected capacity of every head.

\subsection{Parameters and baseline fidelity}
Section~\ref{sec:method} specifies the Twin parameters; only $r$ varies in the main comparison. The allocation reference is AdaKV commit \texttt{04497abac4c1a58426f3daf1014578990e225cc5}. We retain its global-count rule and half-uniform floor, adapting the available budget to protected positions and applying a separate score within each head.

SnapKV uses observation window 64 and pooling width five; PyramidKV uses the same window and kernel with $\beta=20$. StreamingLLM uses four sinks.

Baselines use their original project selection and rounding conventions, including PyramidKV's layer-specific capacities. We do not add \Twin's scoring or protected positions to a baseline. H2O uses the shared SDPA reader with a uniform head budget. Consequently, matching $r$ does not assert identical token identities, layer distributions, or rounding at every individual layer. The exact $HB$ proof applies to \Twin.

\subsection{Additional baseline implementations}
\textbf{KeyDiff.} We use the project's \texttt{KeyDiffPress} scorer: negative cosine similarity between a key and the mean of normalized prefix keys. The native scoring press retains the top-ranked original K/V entries. This is a per-prefix adaptation under the shared evaluation protocol; it does not reproduce the original paper's separate blockwise wrapper.

\textbf{FullKV.} The reference disables compression entirely ($r=0$), while preserving the same context cap, tokenized question, generation limits, and scorer. Its outputs are measured once and reused across displayed ratios. It is not included in compressed-budget rankings.

\textbf{H2O observation rule.}
\label{sec:h2o-implementation}
We use KVPress~\citep{devoto2025expectedattention}, calling its unchanged \texttt{ObservedAttentionPress} scorer and top-$k$/gather selector. It sums causal prefix attention, normalizes each position by its number of visible queries, and averages grouped query heads. This is a prefix-only heavy-hitter adaptation, not the original online eviction policy: no extra sink/recent reservations or decoding-time eviction are added. We materialize weights in bounded multi-head batches, preserving batched GEMM, while retaining the matched SDPA prefix output and reader.

\subsection{Dataset preparation and scoring}
\label{sec:cohort-scoring}
LongBench assigns 13 examples to eight English tasks and 12 to the remaining eight, for 200 examples with a 4,000-token context cap. LooGLE uses 50 examples from each of its four splits, 200 total, with a 16,000-token cap. RULER uses 200 examples at each of 4,096, 8,192, and 16,384 tokens, covering all 13 task types at each length. Each method--model--ratio condition therefore contains 1,000 predictions. The manifests record the exact evaluation subsets.

The six compressed methods and FullKV contribute 76,000 predictions. FullKV is evaluated once and reused across ratios. LongBench weights all 16 task scores equally. LooGLE equally weights its four primary split scores: parsed key-level exact match for short-dependency cloze and the project's ROUGE-L scorer for the remaining splits. RULER averages its 13 task scores at each length, then weights the three lengths equally.

LooGLE short-dependency cloze parses a dictionary and aggregates matches over reference keys; a parsing or dictionary-format failure contributes zero matches and one denominator unit under the fixed scorer. We aggregate the stored task scores, including the original scorer's rounding, with decimal arithmetic. Displayed values are rounded only after aggregation.

The environment records PyTorch 2.9.1, Transformers 4.57.3, Datasets 2.21.0, and pandas 2.3.3. Evaluation uses four NVIDIA L20 GPUs and native SDPA with unchanged model loading defaults and no FP8 quantization. Task-specific generation limits come from the same frozen dataset rows. The chat-template date is fixed to \texttt{08 Sep 2026}; Qwen thinking is disabled, preventing template drift across runs.

\subsection{Implementation checks}
\label{sec:implementation-checks}
Input token hashes are matched across methods, and saved predictions are independently re-scored during final verification. This covers the 16,000 additional KeyDiff and FullKV predictions and the 12,000 new H2O predictions. Raw predictions are saved before calling scorers that may modify their input data frame. For H2O, dense-reference score/selection checks and four-model 16k preflight checks pass.

Controlled Llama and Qwen3 tests confirm bitwise equality of the $r=0.5$ packed caches and checked outputs with the frozen implementation. Uniform-quota native packed and ragged paths agree for question prefilling and a decode step. Tests cover short-prefix protection clipping and exact quotas at all three ratios. Every evaluated \Twin\ prefix records per-layer head counts and K/V bytes, verifying Equation~\ref{eq:budget} without head padding. Model parameter versions and hook cleanup are also checked.

\textbf{Artifacts.} Recorded artifacts include source tables and validation scripts for aggregate and task scores, cross-model controls and all random seeds, sample identifiers, predictions, input-hash evidence, saved configurations, metric records, cache witnesses, and source hashes. The full cost suite and its raw repeats are also recorded; weights and the environment are external dependencies.

\subsection{Stage-efficiency protocol}
\label{sec:efficiency-details}
Figure~\ref{fig:efficiency-stages} uses PyTorch 2.9.1+cu128 on an exclusive NVIDIA L20 with matched native SDPA readers; the host is shared. Its three RULER-4k contexts are the lowest hashes among distinct raw contexts under the frozen selection rule. The exact context and question token hashes match the quality evaluation. All five baselines, FullKV, and Twin are measured in the same extended run; method order rotates across repetitions. Each condition is warmed once and measured three times. Native selection, ordering, and budget rounding are preserved, including PyramidKV's layer-specific capacities.

Within each repeat, TTFT is construction time plus the first read's question-prefill/first-token time; these two phases are added before aggregation. Decode is that read's subsequent 31-step autoregressive duration divided by 31. EOS stopping is disabled only for this fixed-workload timing. We take a median over the three repeats for each context, then average the three context medians. Peak memory is PyTorch's maximum allocated memory over construction and four fresh-suffix reads, including weights and temporaries, not reserved memory or KV payload alone. The repeated reads use the same question and are not a multi-question coverage test.

CUDA synchronization bounds each timing interval. All formal measurements return allocated memory to their starting level. The first warmup initialized a persistent 32\,MiB cuBLAS workspace; a separate cold-start replay and workspace-release check identified this allocation. It is excluded with the other warmups. Measured repeats use the same per-context memory baseline across methods.

\section{Component Experiments and Selection Diagnostics}
\label{sec:new-confirmation}
\label{sec:diagnostics}
\subsection{Score and capacity comparisons}
The RULER component experiments use 200 instances per length stratified over 13 tasks. The LooGLE component experiments use 32 summarization documents.

In the four-cell controls, V/T denote value-only/Twin scoring and U/A denote uniform/adapted capacity. One prefill supplies all cells, with head quotas reused across score variants. Absolute scores follow the task scorer's rounding; bootstrap contrasts use unrounded document contributions.

\begin{table}[htbp]
\centering\small
\caption{\textbf{Score and capacity comparisons on 32 LooGLE summary documents.} FullKV is uncompressed.}
\label{tab:confirmation-all}
\setlength{\tabcolsep}{4pt}
\begin{tabular}{lllrrrrr}
\toprule
Model & Scope & $r$ & V/U & V/A & T/U & T/A & FullKV \\
\midrule
Qwen3-4B & Summary & 0.5 & 31.04 & 32.16 & 30.23 & 30.75 & 31.91 \\
Qwen3-4B & Summary & 0.75 & 29.79 & 30.12 & 28.79 & 26.89 & 31.91 \\
Llama-3.2-1B & Summary & 0.5 & 25.63 & 26.62 & 25.78 & 25.96 & 27.04 \\
Llama-3.2-1B & Summary & 0.75 & 23.18 & 22.00 & 22.47 & 21.84 & 27.04 \\
\bottomrule
\end{tabular}
\end{table}
\begin{table}[htbp]
\centering\small
\caption{\textbf{Summary-branch permutation controls at $r=0.75$.} True denotes the original frequency--position correspondence.}
\label{tab:shuffle-all}
\begin{tabular}{llrrrr}
\toprule
Model & Scope & True & Seed 11 & Seed 29 & Seed 47 \\
\midrule
Qwen3-4B & Summary & 26.89 & 28.01 & 29.57 & 29.45 \\
Llama-3.2-1B & Summary & 21.84 & 23.08 & 24.00 & 22.28 \\
\bottomrule
\end{tabular}
\end{table}
\begin{table}[htbp]
\centering\small
\caption{\textbf{Score and capacity comparisons on LongBench.}}
\label{tab:development-four}
\begin{tabular}{llrrrr}
\toprule
Model & $r$ & V/U & V/A & T/U & T/A \\
\midrule
Qwen3-4B & 0.5 & 39.02 & 38.68 & 38.70 & 37.20 \\
Qwen3-4B & 0.75 & 34.26 & 36.09 & 34.97 & 35.51 \\
Llama-3.1-8B & 0.5 & 36.94 & 36.02 & 37.12 & 36.60 \\
Llama-3.1-8B & 0.75 & 34.41 & 35.11 & 34.69 & 35.04 \\
\bottomrule
\end{tabular}
\end{table}

\subsection{Measured frequency statistics}
Figure~\ref{fig:frequency-distribution} pools the six-bin count histogram over layers and KV heads within each context, excluding protected positions, normalizes it, and averages the 200 context distributions at each length. The discount mean first averages layer/head means within a context. Selection change is the number of newly retained nonprotected entries relative to value-only, divided by the retained nonprotected count; contexts are averaged equally. All statistics use original prefix keys before any future question is exposed.

\paragraph{Selection-boundary distributions.}\label{sec:boundary-distribution}
Figure~\ref{fig:selection-boundary} uses all contexts from the matched-replacement diagnostic for both models: 78 RULER contexts and eight LooGLE summary documents. In each layer/head, the cutoff is the lowest nonprotected value energy retained by value-only at the fixed adapted capacity. We measure $|e_i/e_{\mathrm{cut}}-1|$ for Twin-added positions, Twin-removed positions, and the eligible pool. At most 16 positions per group/head are sampled independently of answers. We weight each by group population divided by sampled count, normalize within a context, and average contexts equally. The five bins cover the entire range, including all deviations above 50\%; no outliers are dropped.

\subsection{Resampling and correspondence controls}
All intervals below are paired 95\% document-bootstrap intervals with 2,000 resamples and no multiplicity correction. The resamples preserve 39 equally weighted RULER task--length cells; summary resampling uses documents. The core-score ablation yields full-minus-value and full-minus-rarity effects of 3.14 points ($[1.34,4.89]$) and 11.43 points ($[8.48,14.59]$). For the Llama-3.2-1B RULER comparison in Table~\ref{tab:correspondence} at $r=0.75$, the adapted and uniform effects are 5.54 ($[3.70,7.41]$) and 5.48 ($[3.40,7.62]$); the true-minus-permuted-mean effect is 7.41 ($[5.47,9.35]$). A secondary analysis groups the 600 RULER instances into 585 exact task/question/reference families and shares each family's resampling multiplicity across lengths. Llama's adapted/uniform frequency-gain intervals are $[3.56,7.60]$/$[3.21,7.61]$, respectively; the true-minus-permuted-mean interval is $[5.55,9.44]$.

At $r=0.75$, summary true-minus-shuffled-mean effects are $-2.12$ for Qwen3-4B ($[-4.02,-0.37]$) and $-1.28$ for Llama-3.2-1B ($[-2.60,-0.01]$); Llama's individual seed 11 and 47 intervals span zero. Llama RULER scores for seeds 11, 29, 47 are 30.97, 29.70, and 30.00. Local-exclusion and discount-shape controls show no consistent cross-model advantage. Relative to true Twin, permutations change approximately 19.60\%/9.76\% of retained nonprotected positions on Qwen/Llama RULER and 15.96\%/7.66\% on summary. Different perturbation magnitudes limit an equal-size intervention interpretation.
The repeated RULER comparison uses another 200 contexts at each of the same three lengths. Parameters, generation, task weights, and document resampling remain fixed. Its Llama-3.2-1B frequency effects are 3.625 points under adapted capacity ($[1.59,5.70]$) and 3.272 under uniform capacity ($[1.28,5.24]$), at $r=0.75$.

\paragraph{Matched replacements.}\label{sec:matched-replacements}
The matched-replacement experiment uses 78 RULER contexts (two per task and length), selected by SHA256 ordering of sample IDs. In each layer/head, let $m$ be the number of value-only positions replaced by Twin. Each control removes and adds $m$ entries without replacement, matching the true changes' joint histogram over ten energy-rank bins and ten position bins. Seeds 11, 29, and 47 are all reported. For Llama RULER, approximately 62.5\% of the control's added/removed members differ from the true replacements; the remaining overlap is retained. The comparison uses the same adapted capacities, protected positions, original K/V, and reader. Its 2,000 paired document resamples preserve all 39 task--length weights; the true-minus-matched-mean effect is 5.93 points with interval $[1.98,9.92]$. Energy-margin statistics sample at most 16 positions from each changed set per head, independently of answers. Each sampled position is weighted by its set size divided by the number sampled, followed by context averaging.

\FloatBarrier

\ifreaderlayout
\clearpage\AIStatement
\ReproStatement
\bibliography{iclr2027_conference}
\bibliographystyle{iclr2027_conference}
\fi
\end{document}